\documentclass[11pt,twocolumn,journal]{IEEEtran}
\usepackage{times}
\usepackage{epsfig}
\usepackage{graphicx}
\usepackage{amsmath}
\usepackage{amssymb}
\usepackage{booktabs}
\usepackage{subfig}
\usepackage{float}
\usepackage{xcolor}

\newcommand{\etc}{\emph{etc.} }

\newcommand{\ie}{\emph{i.e.} }
\newcommand{\vs}{\emph{vs.} }

\newcommand{\fig}{Fig. }
\newcommand{\tab}{Tab. }
\newcommand{\SMPR}{SMPR}

\graphicspath{{figs/}}
\begin{document}

\title{\SMPR: Single-Stage Multi-Person Pose Regression}
%
%



\author{Junqi Lin, Huixin Miao, Junjie Cao*, Zhixun Su, Risheng Liu
\thanks{J. Lin and H. Miao equally contribute to this paper. All the authors are with Dalian University of Technology. 
J. Lin is also with Business Growth BU, JD.com. 
J. Cao and Z. Su are also with Key Laboratory for Computational Mathematics and Data Intelligence of Liaoning Province.}
\thanks{E-mail: jjcao@dlut.edu.cn.}
}


\maketitle

\begin{abstract}
Existing multi-person pose estimators can be roughly divided into two-stage approaches (top-down and bottom-up approaches) and one-stage approaches. The two-stage methods either suffer high computational redundancy for additional person detectors or group keypoints heuristically after predicting all the instance-free keypoints. The recently proposed single-stage methods do not rely on the above two extra stages but have lower performance than the latest bottom-up approaches. In this work, a novel single-stage multi-person pose regression, termed \SMPR, is presented. It follows the paradigm of dense prediction and predicts instance-aware keypoints from every location. Besides feature aggregation, we propose better strategies to define positive pose hypotheses for training which all play an important role in dense pose estimation. The network also learns the scores of estimated poses. The pose scoring strategy further improves the pose estimation performance by prioritizing superior poses during non-maximum suppression (NMS). We show that our method not only outperforms existing single-stage methods and but also be competitive with the latest bottom-up methods, with 70.2 AP and 77.5 AP75 on the COCO test-dev pose benchmark. Code is available at https://github.com/cmdi-dlut/SMPR.

\end{abstract}

\section{Introduction}
Multi-person pose estimation from a single image aims to identify all the person instances and detect the keypoints of each person simultaneously. It is a challenging task in computer vision, and widely used in motion recognition \cite{wang2018rgb, wang2013approach}, person Re-ID \cite{li2018harmonious}, and pedestrian tracking \cite{zhu2013apt}, \etc

Previous neural network methods are mostly two-stage based, which are roughly divided into top-down and bottom-up approaches. 
The top-down methods \cite{chen2018cascaded, gkioxari2014using, fang2017rmpe, rogez2019lcr, sun2011articulated, sun2019deep} employ object detectors to identify all the persons and then detect their keypoints individually. 
The bottom-up methods \cite{cao2017realtime, insafutdinov2016deepercut, newell2017associative, pishchulin2016deepcut, papandreou2018personlab} detect all instance-free keypoints first and then group them into person instances heuristically.
On the one hand, these two-stage strategies achieve higher performance by dividing the challenge into smaller sub problems which are easier to be solved. On the other hand, they impose additional computational overheads for the computation is separated or not fully shared. They also introduce some essential obstacles. For example, the top-down methods' running time heavily depends on the number of persons in the image and there is no clear semantic connection between keypoints when grouping them into instances in the bottom-up methods.

\begin{figure}[t]
\setlength{\belowcaptionskip}{-0.7cm} 
\includegraphics[width=\linewidth, keepaspectratio]{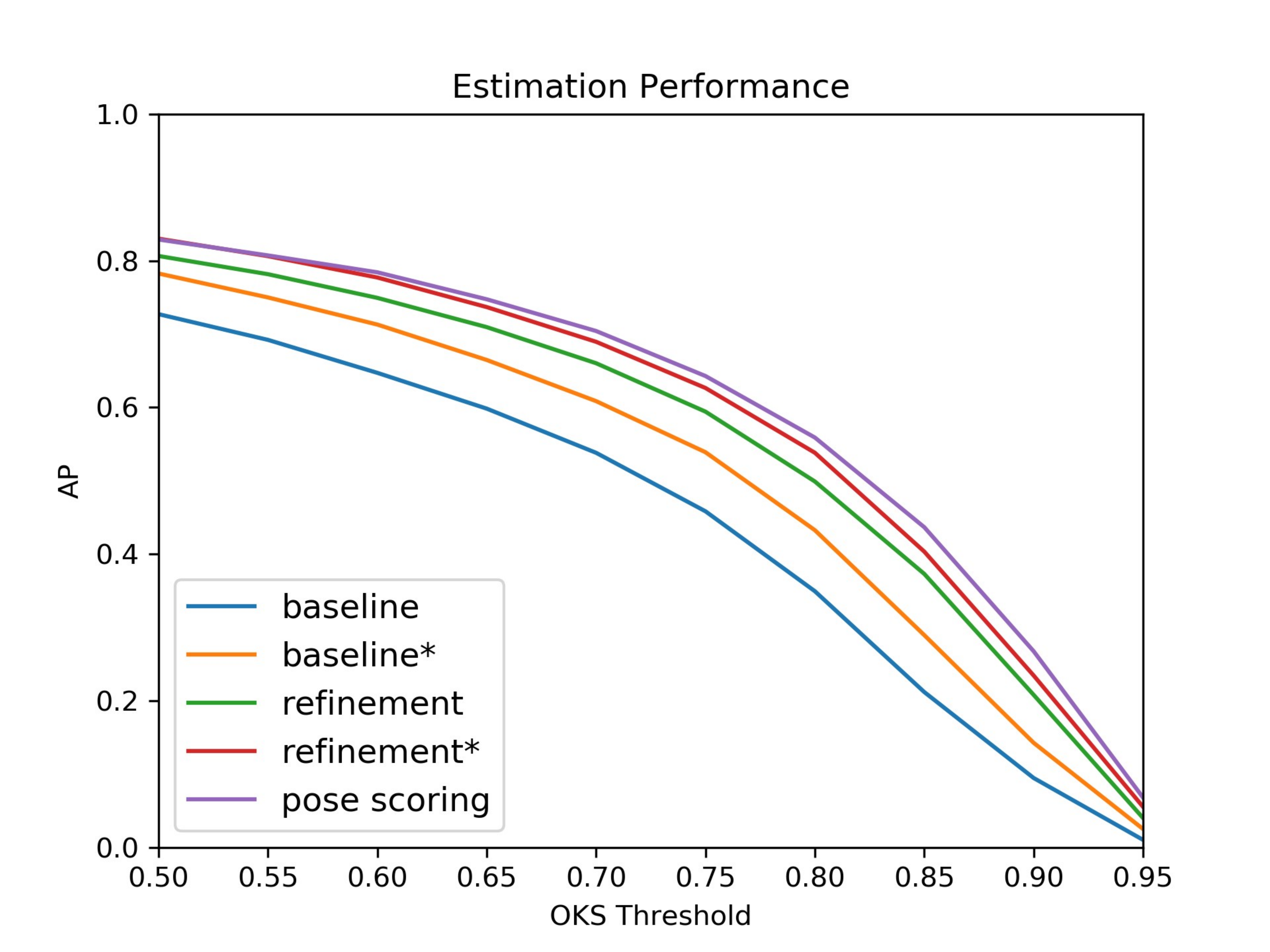}
\caption{Superior pose hypotheses identification
is vital for dense pose regression. "baseline" and "baseline*" means initial pose regression with different positive poses selection strategies. "refinement" and "refinements*" means pose refinement after feature aggregation with different positive poses selection strategies. Only classification score is used in NMS for the above comparisons. "pose scoring" means that predicated pose scores are also utilized in NMS. The comparison are done on COCO minival. 
}
\label{fig_teaser}
\end{figure}

\begin{figure*}[t]
\begin{center}
\begin{tabular}{@{}c@{}}
\includegraphics[width=1\textwidth, keepaspectratio]{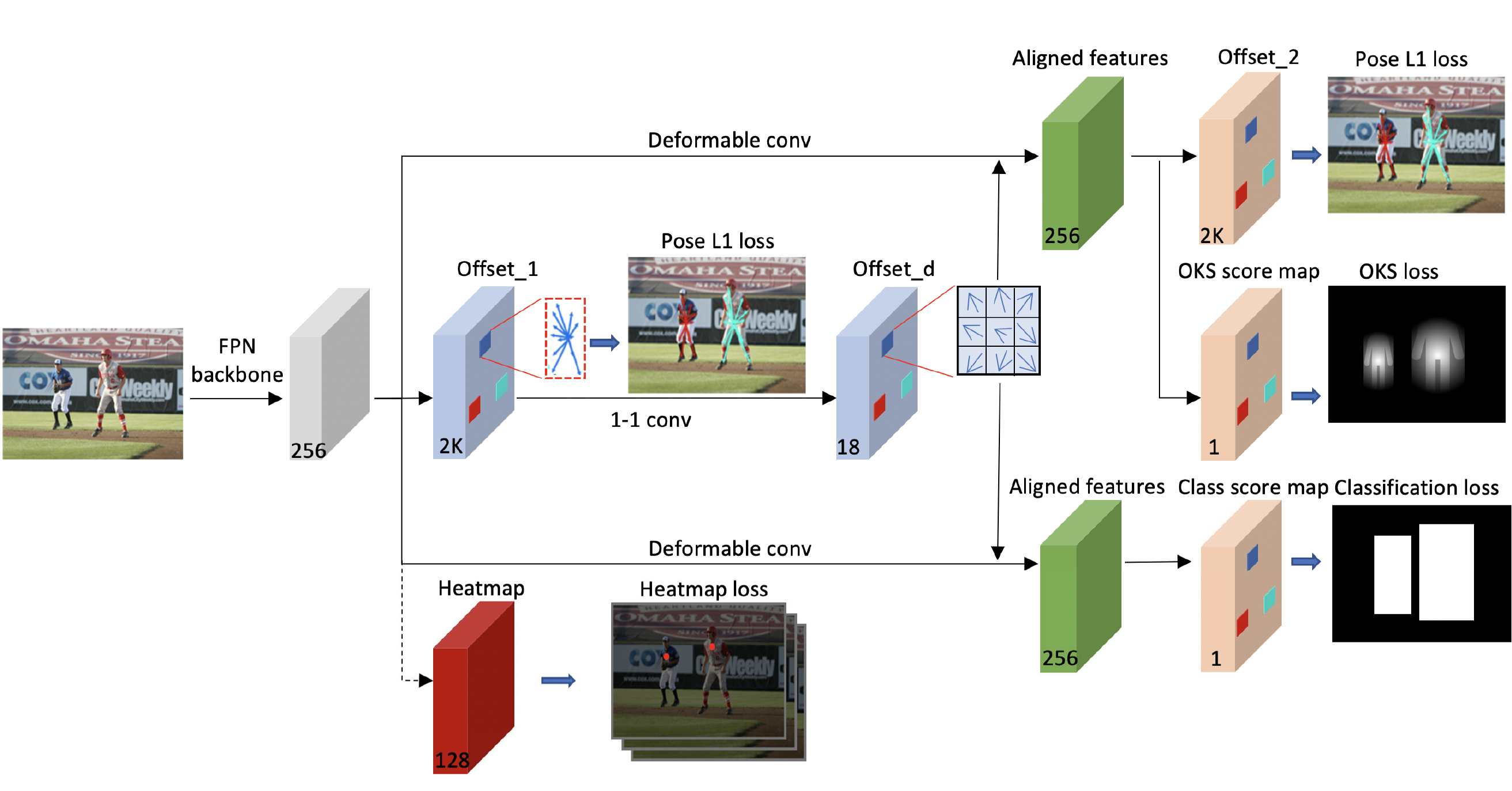}
\end{tabular}
\end{center}
\vspace{-2mm}
\caption{Overview of the proposed \SMPR. \SMPR~adopts a FPN backbone. 
We only draw the afterwards pipeline of one scale of FPN feature maps for clear illustration. There are 5 heads for the feature maps from level 3 to level 7. 
Each head has two subnets: one pose regression and the other for location classification which labels the locations on the feature maps with “person” or “not person”. The regression subnet has three branches aiming for initial pose estimation (location + offsets 1), final pose regression (location + offsets 1 + offsets 2) and scoring. 
In the head for feature level 3, a heatmap prediction subnet is included to assist training.
}
\label{fig_framework}
\end{figure*}

Very recently, a single-stage paradigm is proposed~\cite{spm19, directpose19, pointset20} to overcome the aforementioned limitations. 
These methods estimate all the instance-aware keypoints of an input image in a compact and efficiency manner. 
Single-stage pose estimation seems similar with single-stage object detection~\cite{fcos19,reppoints19}. They all regress $K$ 2D points from each feature location using one single feature vector of that location. 
However predicting $K$ keypoints, $K=17$ in COCO \cite{coco14} dataset, from a feature vector is harder than estimating 2 corners of a bounding box from the feature vector. 
Because keypoints usually not lie in proximity of the location which needs larger receptive field and keypoints own much more geometric variations than box corners. Modules based on hierarchical keypoints representation \cite{spm19} or feature aggregation by deformable convolution (DCN) \cite{directpose19, pointset20} are designed to improve the initial keypoints regression by locating more informative features lying closer to the keypoints. 
It is evident that predicting the keypoints using multiple nearby feature vectors, for example, regressing the wrist keypoint with a feature vector around the wrist, is much easier than using a single feature vector far away from them. 
But they still perform less well than the latest bottom-up methods.

There are two problems in current single-state pose estimators. The first problem is that including false positive poses in regression losses, and the second is that superior pose may be suppressed in NMS using classification scores. 
\fig \ref{fig_teaser} show that positive pose hypothesis selection, feature aggregation and pose scoring are all important for dense pose regression. 
The two problems can actually be improved using one common perspective, i.e. identifying positive/superior poses. Following the perspective, we propose a single-stage multi-person pose regression network in an anchor-free way, called \SMPR. It estimates all the instance-aware keypoints of an input image in a compact and efficiency manner, see \fig \ref{fig_framework}. \SMPR~adopts feature pyramid networks (FPN) \cite{fpn17} and has 5 heads for the feature maps from level 3 (downsampling ratio of 8) to level 7 (downsampling ratio of 128).
Each head contains multiple branches for classification and pose regression, etc. 
We regress one initial pose for each feature location of current feature level and refine the pose after feature aggregation, and the two regressions are both supervised by positive poses.  
We adopt three positive pose identification strategies for the initial pose regression, the final pose regression and the NMS step. Considering that the initial pose regression from single feature location is not trusty, we select a pose as positive if its feature location is inside a shrunk minimal enclosing bounding box of the corresponding pose, which means that the used feature is very close to the box's center. When refining the initial regression, OKS (Object Keypoint Similarity) of the refined poses are adopted to select positives since the refined poses are far superior than the initial estimation. Finally, we also learn a OKS score for NMS in view that the classification score is not very relevant to the quality of a pose. 
Our \SMPR, with the three superior pose identification strategies, outperforms other single-stage pose estimators. It also shows competitive performance with the latest bottom-up pose estimators~\cite{higherhrnet2020}. The contributions of this paper are summarized as follows: 
\begin{itemize}
    \item We propose two positive pose identification strategies during training which improve the performance of single-stage pose estimation.  
    \item We present a pose scoring module to address the problem of sorting pose hypotheses in NMS. It explores a new direction for improving the performance of pose estimators.
    \item We propose an anchor-free single-stage network for multi-person pose estimation, which reduces the computational redundancy. It is end-to-end trainable and does not require additional pose anchors or handcrafted keypoints grouping.
\end{itemize}

\section{Related Work}
\noindent \textbf{Top-down Methods.} \quad In the top-down methods \cite{chen2018cascaded, gkioxari2014using, fang2017rmpe, rogez2019lcr, newell2016stacked, chen2018cascaded, sun2019deep}, person detectors is first employed to generate bounding boxes for each person instance in an image. Then, the regions in the bounding boxes are cropped and scaled to perform single-person pose estimation. Mask RCNN \cite{he2017mask} proposes to utilize extracted features instead of the original image to improve efficiency. In order to focus on regressing difficult keypoints and improve accuracy further, CPN \cite{chen2018cascaded} proposes a cascaded pyramid network. From the perspective of feature encoding, HRNet \cite{sun2019deep} proposes a high-resolution network to maintain high-resolution feature representations, which achieves the state-of-the-art performance. Top-down methods have better performance, but they cannot achieve computing sharing. Moreover, they are highly dependent on the performance of person detectors.

\hspace*{\fill} \\
\noindent \textbf{Bottom-up Methods.} \quad The bottom-up methods \cite{cao2017realtime, insafutdinov2016deepercut, newell2017associative, higherhrnet2020} first detect all instance-free keypoints and then group them together. For grouping, Openpose \cite{cao2017realtime} proposes to utilize part affinity fields to establish connections between keypoints, and then uses the greedy algorithm to combine the corresponding keypoints with the highest scores. Associative Embedding \cite{newell2017associative} proposes to group keypoints by generating tags for each keypoint, and those with the same tag belong to the same person. Based on this grouping strategy, HigherHRNet \cite{higherhrnet2020} proposes to use the fusion of multi-resolution heatmaps to implement keypoints detection. Bottom-up methods can share calculations, so they are faster. But grouping is heuristic, which makes the model difficult to train. Their performance are usually lower than the top-down methods.

\hspace*{\fill} \\
\noindent \textbf{Single-Stage Methods.} \quad 
The single-stage methods \cite{spm19, centernet19, directpose19, pointset20} are proposed very recently to overcome the aforementioned difficulties. The structured pose representation is introduced, which represents the keypoints of each person instance as a root location with $K$ offsets. Hence these methods predict instance-aware keypoints directly from estimated root locations \cite{spm19, centernet19} or follow the dense predication strategy \cite{directpose19, pointset20}, \ie~predicate poses from each location of the feature maps.
To regress a pose from one root location accurately, SPM \cite{spm19} adapts an hierarchical strategy. CenterNet \cite{centernet19} estimates individual keypoints using standard bottom-up approach and assign the individual keypoints to their closest person instance indicated by a root location and $K$ offsets. 
The dense pose estimators \cite{directpose19, pointset20} employ DCN \cite{dcn17} to refine the initial estimation from each location. Point-set anchors \cite{pointset20} provides 27 pose anchors per location as the initial pose estimation and DirectPose \cite{directpose19} and our \SMPR~adopt anchor-free approach. The performance of DirectPose is poor even with a delicate feature aggregation mechanism. Although SPM and point-set anchors outperform many state-of-the-art bottom-up methods, they are not competitive with HigherHRNet \cite{higherhrnet2020}. 
Our \SMPR~outperforms previous one-stage methods by a large margin (more than 7.2 points AP than DirectPose and 1.5 points AP than Point-set Anchors with smaller model size). 


\section{Our Method}
\begin{figure}[H]
\includegraphics[width=\linewidth, keepaspectratio]{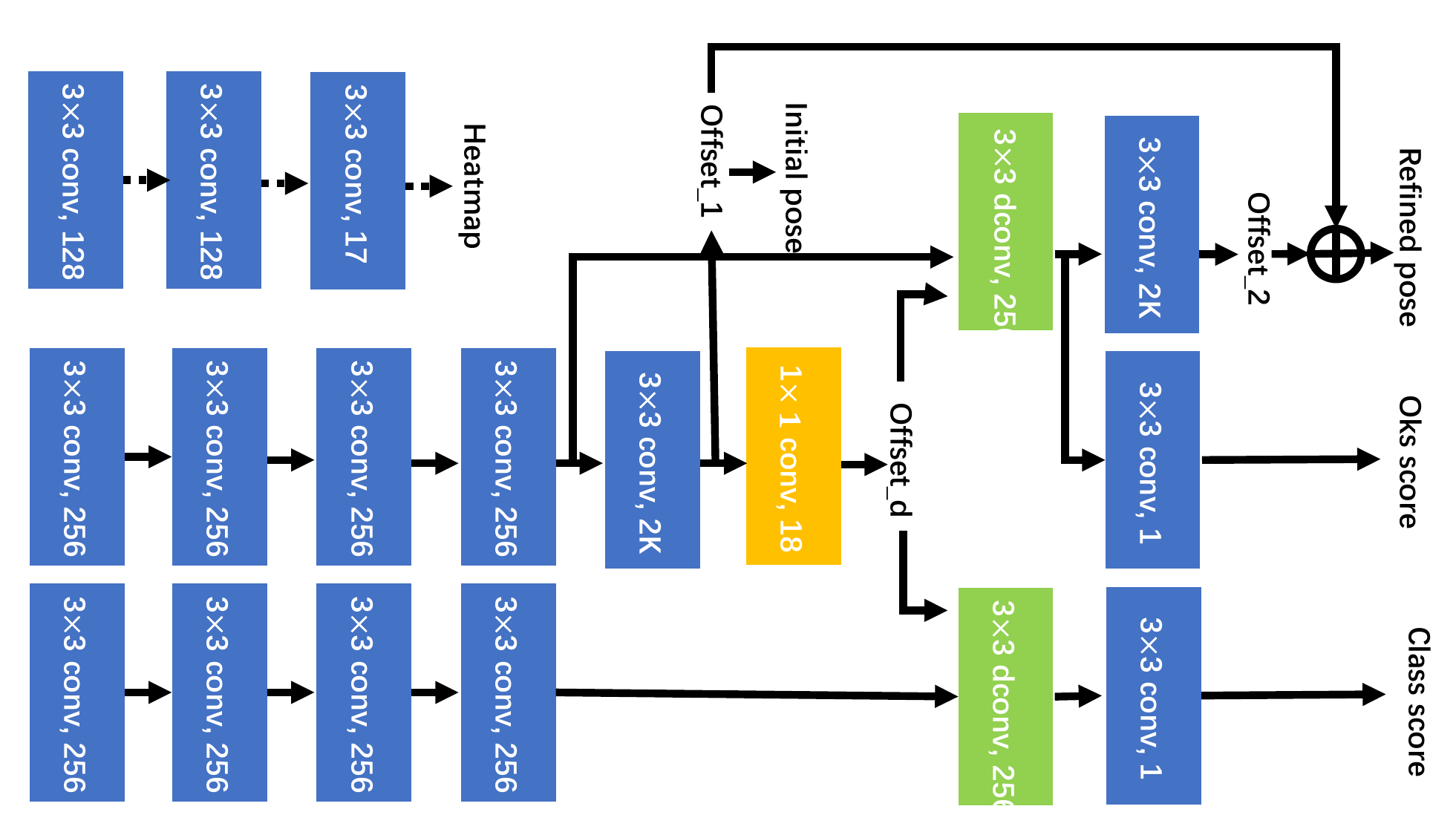}
\caption{The head architecture of \SMPR.}
\label{fig_head}
\end{figure}

In this section, we first introduce the instance-aware keypoints representation. \SMPR~is designed based on the representation and its network architecture are discussed next.
Then we show how to predicate the initial poses, align features with keypoints' position and refine the initial poses using the aligned features. We observe that selection of positive poses is vital for pose regression. Hence two different positive selection strategies are detailed when introducing the initial pose predication and the final pose refinement. 
Finally, we illustrate the proposed pose scoring module (PSM) which solves the inconsistency between classification scores and quality of predicated poses. In each subsection, the relationship to and differences from existing single-stage pose estimators are also discussed.

\subsection{Instance-aware Pose Representation}
Poses are conventionally represented as 
\begin{equation}\label{pose_rep1}
    \bar{\mathcal{P}} = \{P_i^1, P_i^2, ..., P_i^K\}_{i=1}^{N},
\end{equation}
where $N$ is the number of persons in the image, $K$ is the number of keypoints, and $P_i^j=(x_i^j, y_i^j)$ denotes coordinates of the $j$th keypoint from person $i$. 
\SMPR~estimates poses in an image in a dense way, i.e. predicating one pose from each location. We need an 
instance-aware pose representation aims to unify the location information of person instance and keypoints, which can be defined as 
\begin{multline}\label{pose_rep2}
    \mathcal{P} = \{(x_i^c, y_i^c), (\Delta x_i^1, \Delta y_i^1), (\Delta x_i^2, \Delta y_i^2), ..., \\ 
    (\Delta x_i^K, \Delta y_i^K),\}_{i=1}^{N},
\end{multline}
where $(x_i^c, y_i^c)$ represents the coordinates of a location for predicating the initial keypoints of the $i$th person. Then the $j$th keypoint of the $i$th person can be expressed as $(x_i^j,\ y_i^j) = (x_i^c + \Delta x_i^j,\  y_i^c + \Delta y_i^j)$. $\Delta x_i^j$ and $\Delta y_i^j$ represent the offsets. The structured pose representation (SPR) in \cite{spm19} is also offsets based. 
Note that there are two differences between SPR and our representation. In SPR, an auxiliary root joint $(x_i^r, y_i^r)$ is introduced to denote the person position which is learnt in their network, and multiple person may share the same root joint theoretically. In \SMPR, the location $(x_i^c, y_i^c)$ is known and not shared. It is assigned to person $i$ if it locates in the bounding box of that person. If it locates in the overlap of multiple bounding boxes we assign it to the person with the smallest bounding box.  

\subsection{Network Architecture}

\subsection{Initial Pose Regression}
\begin{figure}[t]
    \includegraphics[width=\linewidth, keepaspectratio]{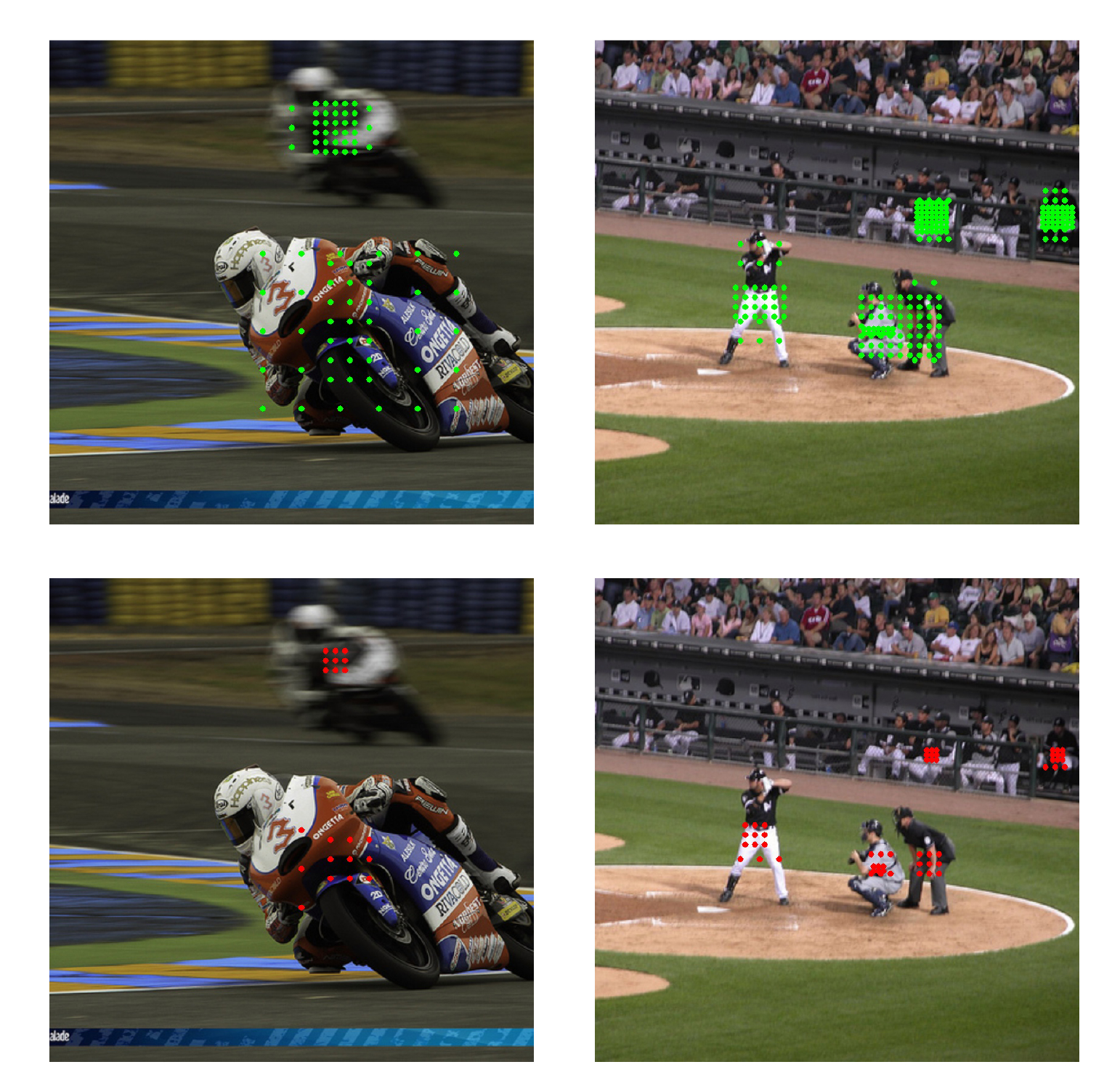}
    \caption{Pose hypotheses from locations inside the pseudo boxes (top row) or shrunk pseudo boxes (bottom row) are selected as positives during the initial pose regression, which leads to 43.3 or 49.5 AP on COCO minival. Locations of positives are represented by green or red dots in the top row or bottom row respectively.}
    \label{fig_position_filter}
\end{figure}

The network architecture of \SMPR~is an extention of the anchor-free object detector RepPoints \cite{reppoints19} with new branches, adapted modules and specified training strategies for pose detection. We use ResNet \cite{he2016deep} or HRNet \cite{sun2019deep} as the backbone network, and then utilize the feature pyramid network (FPN) \cite{fpn17} to produce 5 feature pyramid levels from scale 3 (downsampling ratio of 8) to scale 7 (downsampling ratio of 128).   
Thus there are 5 shared heads for the features of 5 scales. The ground truth poses are assigned to different feature levels according to their scales, and each head is responsible for predicating poses of corresponding scale. 
The head is illustrated in \fig\ref{fig_framework} and its detail architecture can be found in \fig\ref{fig_head}. 
Each head contains two subnets aiming at pose regression and location classification, respectively. Classification labels the locations on the feature maps with “person” or “not person”. 
The regression subnet has three branches aiming for initial pose regression (location + offsets\_1), final pose regression (location + offsets\_1 + offsets\_2) and OKS scoring. 
On each feature location, we estimate an initial pose using the single feature vector of the location. 
Then a feature aggregation module is designed to align the keypoints with more feature vectors located around each keypoint. We predicate the final poses, OKS score and classification score from the aligned features. 
The numbers of output channels of the two regression branches are both 2K (K represents the number of keypoints). 
A heatmap prediction subnet [28] is included only in the head for feature maps from scale 3 to assist training.

We regress one initial pose for each feature location of the current feature level. The initial pose is expressed as $K$ offsets, stored in offsets\_1, plus the feature location. 
The regression is supervised by the ground truth poses. We observe that supervising regression from all the locations reduce the network's performance and only positive samples, i.e. superior poses, should be supervised.  
Superior poses can be measured by OKS. But the initial poses are far from  precise since only one feature vector of current location is used and the location is distant from the predicated keypoints. Hence we prefer to 
identify positives by the relative position of current locations and the nearest pseudo bounding box, which is the minimal bounding box of ground truth keypoints. Specifically, if a location is inside a shrunk area of a pseudo box, we take the predication from it as a positive. The side length of the shrunk boxes for different feature levels are $(8, 16, 32, 64, 128) * 1.5$ respectively. 
Previous single-stage estimators \cite{spm19, directpose19, pointset20} actually all have a step for the initial keypoints predication. \cite{spm19} predicates partial keypoints closer to the root joint. \cite{directpose19, pointset20} and we adopt similar strategies. But \cite{pointset20} employs 27 pose point-set anchors per location and we depend on anchor-free regression. \cite{directpose19} also follows the anchor-free way but the initial predication is not supervised. We find that the supervised regression using positives determined by the shrunk pseudo boxes yields to higher performance than using positives determined by the pseudo boxes, see \fig\ref{fig_position_filter}.

\subsection{Feature Aggregation}
\begin{figure}[t]
    \centering
    \includegraphics[scale=2]{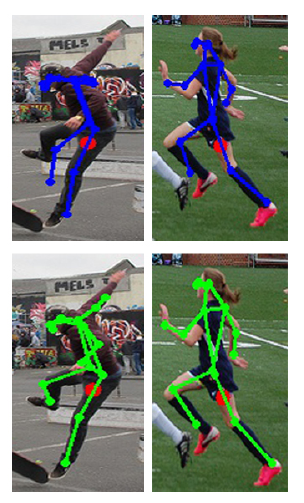}
    \caption{Comparison of poses predicated from the same locations before and after feature aggregation.}
    \label{fig_feature_alignment}
\end{figure}
The initial predicated poses are inferior, see top row of \fig\ref{fig_feature_alignment}. Because it is difficult to decode all the keypoints of a person through a single feature vector of current location, especially for the keypoints far away from the receptive field of the location. 
DCN is employed in \cite{reppoints19, directpose19, pointset20} to learn shape invariant features, i.e. extract more feature vectors closer to the keypoints. 
In principle, superior poses can be decoded from these features. 
We follow the simple strategy of \cite{reppoints19, pointset20}. The initial keypoints field of 34 channels are converted to a offset field, offset\_d in \fig\ref{fig_head}, of 18 channels by 1x1 convolution. Feature aggregation is thus implemented by 3x3 DCN with the specified offset field. Then superior poses can be learnt from the same location using more aligned feature vectors as shown in bottom row of \fig\ref{fig_feature_alignment}. We also utilize the operation to enhance the classification features simultaneously. 

\subsection{Pose Refinement}

\begin{figure}[t]
    \includegraphics[width=\linewidth, keepaspectratio]{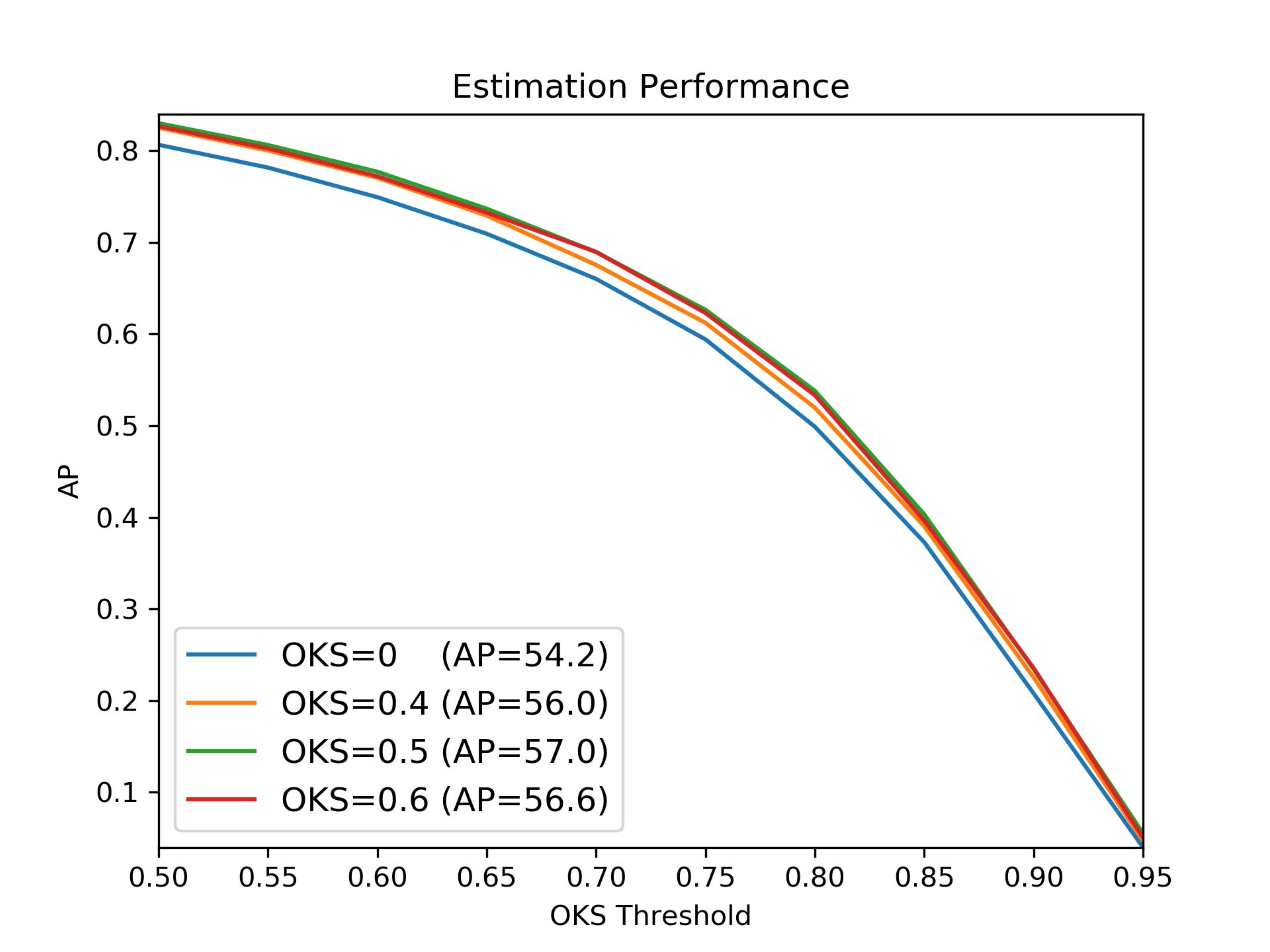}
    \caption{Performance of pose refinement using positives selected by  different OKS thresholds on COCO minival.}
    \label{fig_oks_filter}
\end{figure}

We can regress superior poses based on the aligned features. 
As shown in \fig\ref{fig_head}, the $j$th keypoint of person $i$ is expressed as
\begin{multline}\label{pose_rep3}
    (x_i^j, y_i^j) = (x_i^c + \Delta x_i^j + \bar{\Delta}x_i^j,\ y_i^c + \Delta y_i^j + \bar{\Delta}y_i^j),
\end{multline}
where $\Delta$ and $\bar{\Delta}$ are stored in the predicated offsets\_1 and offsets\_2.
Although the refined poses are superior, compared with the initial poses, there are still many inferior poses. 
As stated above, identifying positives is vital for supervised pose regression. 
We compute the OKS of all the refined poses and the poses with OKS higher than a threshold are selected as positives. The threshold is set to 0.5 empirically, see \fig\ref{fig_oks_filter}.
The poses with lower OKS are filtered out. OKS is defined in \cite{coco14} as: 
\begin{equation}
OKS =  \frac {\sum_i e^{ \frac {-d_i^2}{2s^2\kappa_i^2} }\delta(v_i>0)}{\sum_i [\delta(v_i>0)]},
\end{equation}
where $d_i$ is the Euclidean distance between each ground-truth and predicted keypoint, $v_i$ is the visibility flag of the keypoint, $s$ is the object scale and $\kappa_i$ is a per-keypoint constant that controls falloff. 
It is proved by experiments that the performance is improved by a large margin.

\subsection{Pose Scoring}\label{psm}
\begin{figure}[t]
    \setlength{\belowcaptionskip}{-0.5cm}
    \centering
    \includegraphics[width=\linewidth, keepaspectratio]{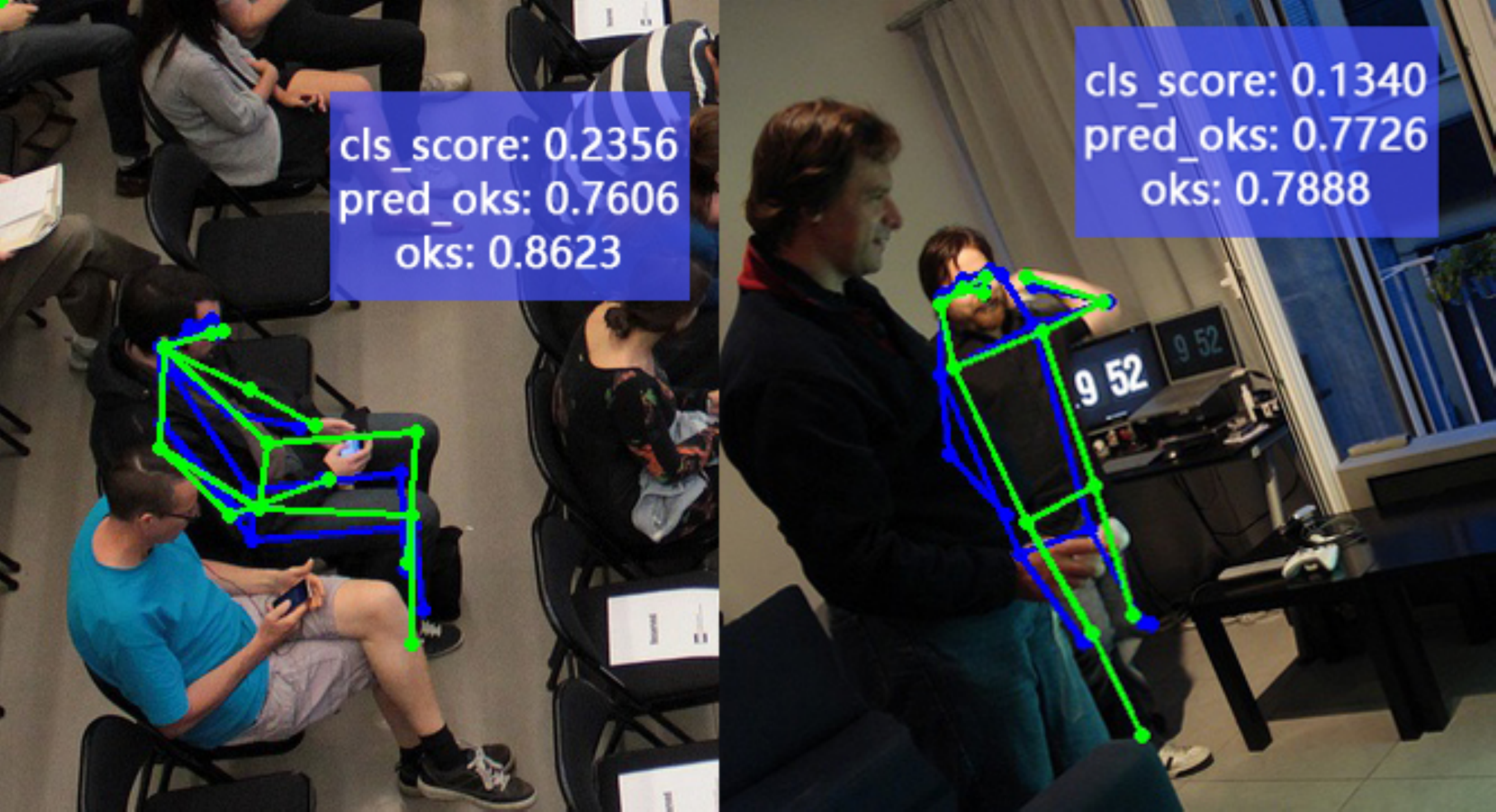}
    \caption{Examples of the inconsistency between classification scores and quality of predicated poses, \ie OKS. The predicated OKS is more correlated with the poses' quality. The green and blue poses represents the ground truth and predicted poses respectively.
    }
    \label{fig:vis_PSM}
\end{figure}

\begin{figure*}[t]
\includegraphics[width=1\textwidth, keepaspectratio]{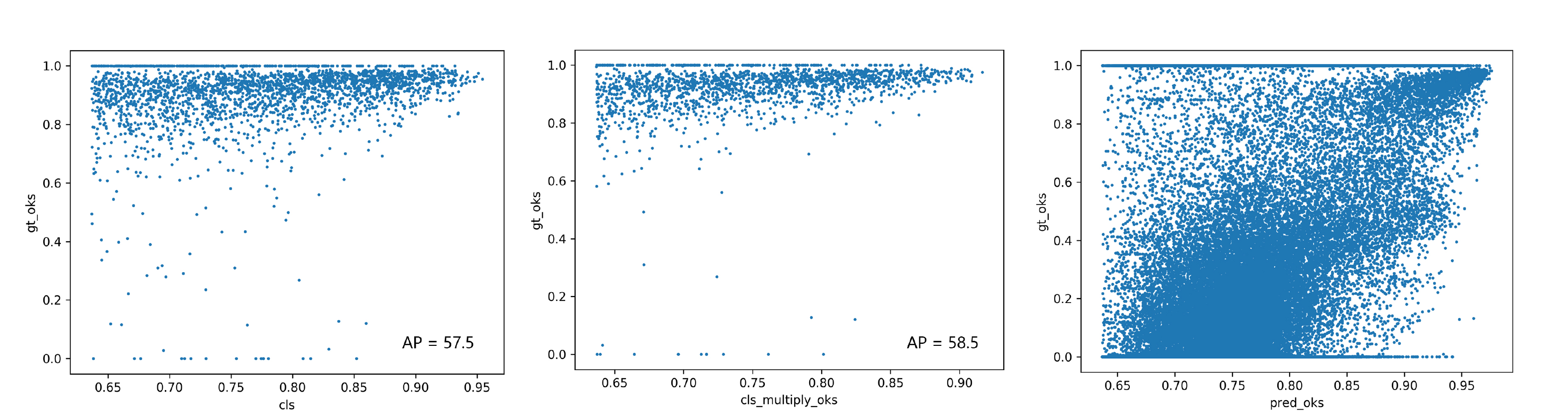}
\vspace{-6mm}
    \caption{Visualizations of predicated classification scores (left), predicated OKS (right) and their multiplication (middle) vs the ground truth OKS.}
    \label{fig:scatter}
\end{figure*}

After the final pose regression, the predictions from all feature levels are merged and NMS is employed as post-processing. Usually, the classification score is used in NMS to select the best poses from the dense estimation. However, the pose quality, measured by OKS, is usually not well correlated with classification score, see \fig \ref{fig:vis_PSM} and \ref{fig:scatter}. Huang \etc \cite{maskScore2019} also meet similar case in instance segmentation. The performance is improved by learning the quality of the predicted instance masks. 
Hence we design a pose scoring module (PSM) to learn the quality of the predicated poses. It is a branch of the pose regression subnet. 
The predicated pose scores are more correlated with OKS, see \fig \ref{fig:vis_PSM} and \ref{fig:scatter}.  
In NMS, the product of the classification score and the pose score is used as the confidence of each final pose. Then superior poses are preserved. Experimental results show that the performance can be improved with PSM.

\subsection{Loss Functions} 
The loss function of the entire network is as follows:
\begin{multline}
    L = \lambda_1 L_{cls} + \lambda_2 L_{hm} + \lambda_3 L_{reg_{initial}} +\\
    \lambda_4 L_{reg_{refined}} + \lambda_5 L_{PSM}, 
\end{multline}
where $\lambda_1 = \lambda_5 = 1$, $\lambda_2 = 4$, $\lambda_3 = 0.05$ and $\lambda_4 = 0.1$. $L_{reg_{*}}$ means L1 loss for the regression branch and $L_{PSM}$ is the binary cross entropy (BCE) loss for PSM. 
Since the training samples on each feature map are mostly negative samples, there is a serious imbalance problem. $L_{cls}$ and $L_{hm}$ are both focal loss functions \cite{lin2017focal} for the classification branch and the heatmap branch, respectively.



\section{Experiments}
\subsection{Implementation}
We present experimental results on the large-scale benchmark COCO dataset \cite{coco14}. Following the common practice \cite{spm19, directpose19, pointset20}, we use trainval35k split (57k images) for training, minival split (5k images) for our ablation studies and test-dev split (20k images) to report our main results. The Average Precision (AP) based on OKS is used as metric. When testing, NMS \cite{bodla2017soft} with a threshold of 0.3 is employed.


Unless specified, ablation studies are conducted with ResNet-50 \cite{he2016deep} and FPN \cite{fpn17}. The network is trained with the stochastic gradient descent (SGD) optimizer over 4 GPU with a mini-batch of 32 images. The initial learning rate is set to 0.02. Weight decay and momentum are set as 0.0001 and 0.9, respectively. Specifically, the model is trained for 30 epochs and the initial learning rate is divided by 10 at epochs 25 and 28. We initialize the backbone network with the weights pre-trained on ImageNet \cite{deng2009imagenet}, and initialize the newly added layers as \cite{lin2017focal}. When training, the input image is resized to have a shorter side of 800 and a longer side less or equal to 1333, and then it is randomly horizontally flipped with probability being 0.5. Finally, it is randomly cropped into 800 x 800 patches. 


\subsection{Ablation Experiments}
\begin{table}[t]
\centering
\begin{tabular}{p{0.2\linewidth}p{0.07\linewidth}p{0.07\linewidth} p{0.07\linewidth}p{0.07\linewidth} p{0.07\linewidth}p{0.07\linewidth}}
\toprule
 & AP & $AP^{50}$ & $AP^{75}$ & $AP^{M}$ & $AP^{L}$ & AR \\
\midrule
baseline & 43.3 & 72.7 & 45.8 & 38.6 & 51.0 & 54.7\\
baseline* & 49.5 & 78.3 & 53.9 & 44.3 & 57.3 & 59.2\\
+ refine & 54.2 & 80.6 & 59.4 & 48.3 & 62.9 & 63.2\\
+ refine* & 57.0 & 83.0 & 62.6 & 50.8 & 65.8 & 64.1\\
+ PSM & 58.5 & 82.9 & 64.3 & 52.3 & 67.1 & 64.8\\
+ heatmap & 60.8 & 84.5 & 67.3 & 54.4 & 69.9 & 67.0\\
\scriptsize - FPN + PAFPN  & 61.6 & 85.0 & 67.1 & 55.6 & 70.2 & 67.7\\
\bottomrule
\end{tabular}
\caption{Ablation experiments on COCO minival. "baseline" and "baseline*" mean training initial pose regression using the samples inside the pseudo bounding box or the shrunk box. 
"+ refine" and "+ refine*" correspond to refine the initial poses after feature aggregation with all or superior pose hypotheses as positives.
"+ PSM" means that predicated pose scores are also used in NMS. 
"+ heatmap" means using heatmap to assist training. 
'- FPN + PAFPN' means using PAFPN instead of FPN to enhance communication between features of different scales.}
\setlength{\belowdisplayskip}{0pt}
\label{tab_ablation}
\end{table}

\subsubsection{Baseline}
First, we conduct the experiments through regressing the initial poses directly to implement multi-person pose estimation, where the samples in the pseudo bounding box are defined as positives. As shown in \tab~\ref{tab_ablation}, the performance is poor (43.3 AP). Because it is hard to regress $K$ keypoints from a single feature vector, especially when its location is far away from the center of the person. Hence the performance is improved when only the samples inside the shrunk pseudo boxes are taken as positives (from 43.3 AP to 49.5 AP), which is denoted as "baseline*". 

\subsubsection{Refinement}
As shown in \tab\ref{tab_ablation}, there is a gap between $AP^{50}$ and $AP^{75}$ of "baseline*", which means that the initial estimation is not accurate but acceptable when measured by a lower OKS. To overcome this issue, feature aggregation is utilized to collect more features close to the keypoints to refine the initial estimation. Then the performance is improved by a large margin (from 49.5 AP to 54.2 AP). But the model has higher AR and relatively low AP, 63.2 \vs~54.2, which means that it can detect person in the image, but it is still difficult to regress keypoints' precise locations. Therefore, we use a OKS threshold, set to 0.5 in this paper, to make the network focus on refine superior initial pose hypotheses. It further improves the performance from 54.2 AP to 57.0 AP.

\subsubsection{PSM}
\begin{table}[t]
\centering
\begin{tabular}{llllll}
\toprule
 & AP & $AP^{50}$ & $AP^{75}$ & $AP^{M}$ & $AP^{L}$ \\
\midrule
+ PSM      & 58.5 & 82.9 & 64.3 & 52.3 & 67.1 \\
Gt Scoring & 68.8 & 87.9 & 75.7 & 64.5 & 75.4 \\
\bottomrule
\end{tabular}
\caption{The upper limit experiment for PSM. "+ PSM" and "Gt Scoring" means using predicated OKS or ground truth OKS in NMS.}
\label{tab_PSM}
\end{table}



As described before, pose quality is usually not well correlated with classification scores. Hence we design PSM to score the predicated poses. Using the product of the classification score and the predicted pose score as the confidence in NMS to improve the performance from 57.0 AP to 58.5 AP. 
We also conduct a upper limit experiment. Using the ground truth OKS as the confidence of the predicated poses in NMS yields to higher performance, 68.8 AP in \tab\ref{tab_PSM}. It shows that the performance of multi-person pose estimators can be improved by a very large margin by designing a more powerful pose scoring module.

\subsubsection{Additional Discussion}
As shown in \tab\ref{tab_ablation}, after using the heatmap with downsampling ratio being 8 (i.e. from scale 3) for auxiliary supervision during training \cite{directpose19}, the performance is improved by 2.3 AP. Note that the heatmap predication is not used during inference. In order to boost information flow and enhance communication between features of different scales, we replaced FPN \cite{fpn17} with PAFPN \cite{liu2018path} to obtain slightly better performance (from 60.8 AP to 61.6 AP).

\subsection{Comparison with State-of-the-art Methods}
\begin{table*}[t]
\begin{center}
\begin{tabular}{ccccccc}
\toprule
Method & Backbone & AP & $AP^{50}$ & $AP^{75}$ & $AP^{M}$ & $AP^{L}$  \\
 \midrule
 \multicolumn{7}{c}{\textbf{Top-down Methods}} \\
 \midrule
 Mask-RCNN \cite{he2017mask} & ResNet-50 & 62.7 & 87.0 & 68.4 & 57.4 & 71.1 \\
 HRNet \cite{sun2019deep} & HRNet-w48 & 75.5 & 92.5 & 83.3 & 71.9 & 81.5 \\
 \midrule
 \multicolumn{7}{c}{\textbf{Bottom-up Methods}} \\
 \midrule
 CMU-Pose \cite{cao2017realtime} & 3CM-3PAF (102) & 61.8 & 84.9 & 67.5 & 57.1 & 68.2 \\
 AE \cite{newell2017associative} & Hourglass-4 stacked* & 63.0 & 85.7 & 68.9 & 58.0 & 70.4 \\
 PersonLab \cite{papandreou2018personlab} & ResNet-152 & 66.5 & 88.0 & 72.6 & 62.4 & 72.3 \\
 HigherHRNet \cite{higherhrnet2020} & HRNet-w32 & 66.4 & 87.5 & 72.8 & 61.2 & 74.2 \\
 HigherHRNet \cite{higherhrnet2020} & HRNet-w48* & 70.5 & 89.3 & 77.2 & 66.6 & 75.8 \\
\midrule
\multicolumn{7}{c}{\textbf{Single-stage Methods}} \\
\midrule
 DirectPose \cite{directpose19} & ResNet-50 & 62.2 & 86.4 & 68.2 & 56.7 & 69.8 \\
 DirectPose \cite{directpose19} & ResNet-50* & 63.0 & 86.8 & 69.3 & 59.1 & 69.3 \\
 CenterNet \cite{centernet19} & Hourglass-2 stacked (104) & 63.0 & 86.8 & 69.6 & 58.9 & 70.4 \\
 SPM \cite{spm19} & Hourglass-8 stacked & 66.9 & 88.5 & 72.9 & 62.6 & 73.1\\
 Point-Set Anchors \cite{pointset20} & HRNet-w48* & 68.7 & \textbf{89.9} & 76.3 & 64.8 & 75.3 \\
 \midrule
 Ours & ResNet-50 & 64.7 & 87.3 & 71.2 & 59.1 & 72.7\\
 Ours & ResNet-50* & 67.1 & 88.5 & 74.2 & 62.0 & 74.7\\
 Ours & HRNet-w32 & 68.2 & 88.7 & 75.3 & 63.3 & 75.4 \\
 Ours & HRNet-w32* & \textbf{70.2} & 89.7 & \textbf{77.5} & \textbf{65.9} & \textbf{77.2} \\
\bottomrule
\end{tabular}
\end{center}
\setlength{\abovecaptionskip}{-0.1cm}
\caption{Comparison with state-of-the-art methods on MS COCO test-dev dataset. "*": using multi-scale testing. 
}
\label{tab_compare}
\end{table*}

We compare \SMPR~with other state-of-the-art multi-person pose estimators on test-dev split of COCO benchmark, see \tab\ref{tab_compare}. We increase the number of epochs from 30 to 100 and divide the initial learning rate by 10 at epochs 80 and 90. With multi-scale testing, \SMPR~achieve 67.1 AP and 70.2 AP with ResNet-50 and HRNet-w32, respectively.

\subsubsection{Comparison with Top-down Methods}
\SMPR~outperforms the classic top-down method -- Mask-RCNN (64.7 AP \vs 62.7 AP). Our model is still behind some latest top-down methods, since they employ an additional person detector to identify people and resolve the inconsistency of human scales. But if the detected boxes overlap, the calculation is redundant and slow. However, the inference time of \SMPR~is independent of the number of person in the input image.

\subsubsection{Comparison with Bottom-up Methods}
\SMPR~outperforms the state-of-the-art bottom-up methods, such as CMU-Pose, AE and PersonLab. 
It also achieves better performance than the latest bottom-up method HigherHRNet \cite{higherhrnet2020}, 68.2 \vs 66.4 AP, using the same backbone HRNet-w32 in single-scale testing. Besides that, our performance with HRNet-w32 is also competitive with HigherHRNet with HRNet-w48 in multi-scale testing. 

\subsubsection{Comparison with Single-stage Methods}
\SMPR~performs far better than most of the single-stage methods, such as DirectPose \cite{directpose19}, CenterNet \cite{centernet19} and SPM \cite{spm19}. 
\SMPR~with the backbone HRNet-w32 even outperforms the latest single-stage pose estimator \cite{pointset20} with the backbone HRNet-w48, 70.2 \vs 68.7 AP, in multi-scale testing. Note that \SMPR~is anchor free and Point-Set Anchors need 27 carefully chosen pose anchors per location.


\section{Conclusion}
In this paper, we present a single-stage multi-person regression network. The proposed network is anchor-free and end-to-end trainable. It estimates multiple instance-aware keypoints in constant inference time. 
Two positive identification strategies are proposed to improve the network's performance. We also learn to score the predicted poses to boost the performance further. 
Experiments demonstrate that the proposed method outperforms many bottom-up and top-down methods, and all the single-stage pose estimators. It achieves 70.2 AP and 89.7 AP50 on the COCO test-dev pose benchmark, which is even competitive with the latest bottom-up methods. 

One interesting future direction would be to improve the pose scoring module. It would also be interesting to identify positives in a differentiable approach.

{\small
\bibliographystyle{ieee}
\bibliography{myref}

\begin{thebibliography}{10}\itemsep=-1pt

\bibitem{bodla2017soft}
N.~Bodla, B.~Singh, R.~Chellappa, and L.~S. Davis.
\newblock Soft-nms--improving object detection with one line of code.
\newblock In {\em ICCV}, pages 5561--5569, 2017.

\bibitem{cao2017realtime}
Z.~Cao, T.~Simon, S.-E. Wei, and Y.~Sheikh.
\newblock Realtime multi-person 2d pose estimation using part affinity fields.
\newblock In {\em CVPR}, pages 7291--7299, 2017.

\bibitem{chen2018cascaded}
Y.~Chen, Z.~Wang, Y.~Peng, Z.~Zhang, G.~Yu, and J.~Sun.
\newblock Cascaded pyramid network for multi-person pose estimation.
\newblock In {\em CVPR}, pages 7103--7112, 2018.

\bibitem{higherhrnet2020}
B.~Cheng, B.~Xiao, J.~Wang, H.~Shi, T.~S. Huang, and L.~Zhang.
\newblock Higherhrnet: Scale-aware representation learning for bottom-up human
  pose estimation.
\newblock In {\em CVPR}, pages 5386--5395, 2020.

\bibitem{dcn17}
J.~Dai, H.~Qi, Y.~Xiong, Y.~Li, G.~Zhang, H.~Hu, and Y.~Wei.
\newblock Deformable convolutional networks.
\newblock In {\em ICCV}, pages 764--773, 2017.

\bibitem{deng2009imagenet}
J.~Deng, W.~Dong, R.~Socher, L.-J. Li, K.~Li, and L.~Fei-Fei.
\newblock Imagenet: A large-scale hierarchical image database.
\newblock In {\em CVPR}, pages 248--255. Ieee, 2009.

\bibitem{fang2017rmpe}
H.-S. Fang, S.~Xie, Y.-W. Tai, and C.~Lu.
\newblock Rmpe: Regional multi-person pose estimation.
\newblock In {\em ICCV}, pages 2334--2343, 2017.

\bibitem{gkioxari2014using}
G.~Gkioxari, B.~Hariharan, R.~Girshick, and J.~Malik.
\newblock Using k-poselets for detecting people and localizing their keypoints.
\newblock In {\em CVPR}, pages 3582--3589, 2014.

\bibitem{he2017mask}
K.~He, G.~Gkioxari, P.~Doll{\'a}r, and R.~Girshick.
\newblock Mask r-cnn.
\newblock In {\em ICCV}, pages 2961--2969, 2017.

\bibitem{he2016deep}
K.~He, X.~Zhang, S.~Ren, and J.~Sun.
\newblock Deep residual learning for image recognition.
\newblock In {\em CVPR}, pages 770--778, 2016.

\bibitem{maskScore2019}
Z.~Huang, L.~Huang, Y.~Gong, C.~Huang, and X.~Wang.
\newblock Mask scoring r-cnn.
\newblock In {\em CVPR}, pages 6409--6418, 2019.

\bibitem{insafutdinov2016deepercut}
E.~Insafutdinov, L.~Pishchulin, B.~Andres, M.~Andriluka, and B.~Schiele.
\newblock Deepercut: A deeper, stronger, and faster multi-person pose
  estimation model.
\newblock In {\em ECCV}, pages 34--50, 2016.

\bibitem{li2018harmonious}
W.~Li, X.~Zhu, and S.~Gong.
\newblock Harmonious attention network for person re-identification.
\newblock In {\em CVPR}, pages 2285--2294, 2018.

\bibitem{fpn17}
T.-Y. Lin, P.~Doll{\'a}r, R.~Girshick, K.~He, B.~Hariharan, and S.~Belongie.
\newblock Feature pyramid networks for object detection.
\newblock In {\em CVPR}, pages 2117--2125, 2017.

\bibitem{lin2017focal}
T.-Y. Lin, P.~Goyal, R.~Girshick, K.~He, and P.~Doll{\'a}r.
\newblock Focal loss for dense object detection.
\newblock In {\em ICCV}, pages 2980--2988, 2017.

\bibitem{coco14}
T.-Y. Lin, M.~Maire, S.~Belongie, J.~Hays, P.~Perona, D.~Ramanan,
  P.~Doll{\'a}r, and C.~L. Zitnick.
\newblock Microsoft coco: Common objects in context.
\newblock In {\em ECCV}, pages 740--755, 2014.

\bibitem{liu2018path}
S.~Liu, L.~Qi, H.~Qin, J.~Shi, and J.~Jia.
\newblock Path aggregation network for instance segmentation.
\newblock In {\em CVPR}, pages 8759--8768, 2018.

\bibitem{newell2017associative}
A.~Newell, Z.~Huang, and J.~Deng.
\newblock Associative embedding: End-to-end learning for joint detection and
  grouping.
\newblock In {\em NeurIPS}, pages 2277--2287, 2017.

\bibitem{newell2016stacked}
A.~Newell, K.~Yang, and J.~Deng.
\newblock Stacked hourglass networks for human pose estimation.
\newblock In {\em ECCV}, pages 483--499, 2016.

\bibitem{spm19}
X.~Nie, J.~Feng, J.~Zhang, and S.~Yan.
\newblock Single-stage multi-person pose machines.
\newblock In {\em ICCV}, pages 6951--6960, 2019.

\bibitem{papandreou2018personlab}
G.~Papandreou, T.~Zhu, L.-C. Chen, S.~Gidaris, J.~Tompson, and K.~Murphy.
\newblock Personlab: Person pose estimation and instance segmentation with a
  bottom-up, part-based, geometric embedding model.
\newblock In {\em ECCV}, pages 269--286, 2018.

\bibitem{pishchulin2016deepcut}
L.~Pishchulin, E.~Insafutdinov, S.~Tang, B.~Andres, M.~Andriluka, P.~V. Gehler,
  and B.~Schiele.
\newblock Deepcut: Joint subset partition and labeling for multi person pose
  estimation.
\newblock In {\em CVPR}, pages 4929--4937, 2016.

\bibitem{rogez2019lcr}
G.~Rogez, P.~Weinzaepfel, and C.~Schmid.
\newblock Lcr-net++: Multi-person 2d and 3d pose detection in natural images.
\newblock {\em TPAMI}, 2019.

\bibitem{sun2019deep}
K.~Sun, B.~Xiao, D.~Liu, and J.~Wang.
\newblock Deep high-resolution representation learning for human pose
  estimation.
\newblock In {\em CVPR}, pages 5693--5703, 2019.

\bibitem{sun2011articulated}
M.~Sun and S.~Savarese.
\newblock Articulated part-based model for joint object detection and pose
  estimation.
\newblock In {\em ICCV}, pages 723--730. IEEE, 2011.

\bibitem{directpose19}
Z.~Tian, H.~Chen, and C.~Shen.
\newblock Directpose: Direct end-to-end multi-person pose estimation.
\newblock {\em arXiv preprint arXiv:1911.07451}, 2019.

\bibitem{fcos19}
Z.~Tian, C.~Shen, H.~Chen, and T.~He.
\newblock Fcos: Fully convolutional one-stage object detection.
\newblock In {\em ICCV}, pages 9627--9636, 2019.

\bibitem{wang2013approach}
C.~Wang, Y.~Wang, and A.~L. Yuille.
\newblock An approach to pose-based action recognition.
\newblock In {\em CVPR}, pages 915--922, 2013.

\bibitem{wang2018rgb}
P.~Wang, W.~Li, P.~Ogunbona, J.~Wan, and S.~Escalera.
\newblock Rgb-d-based human motion recognition with deep learning: A survey.
\newblock {\em CVIU}, 171:118--139, 2018.

\bibitem{pointset20}
F.~Wei, X.~Sun, H.~Li, J.~Wang, and S.~Lin.
\newblock Point-set anchors for object detection, instance segmentation and
  pose estimation.
\newblock In {\em ECCV}, 2020.

\bibitem{reppoints19}
Z.~Yang, S.~Liu, H.~Hu, L.~Wang, and S.~Lin.
\newblock Reppoints: Point set representation for object detection.
\newblock In {\em ICCV}, pages 9657--9666, 2019.

\bibitem{centernet19}
X.~Zhou, D.~Wang, and P.~Kr{\"a}henb{\"u}hl.
\newblock Objects as points.
\newblock In {\em arXiv preprint arXiv:1904.07850}, 2019.

\bibitem{zhu2013apt}
X.~Zhu, Q.~Li, and G.~Chen.
\newblock Apt: Accurate outdoor pedestrian tracking with smartphones.
\newblock In {\em IEEE INFOCOM}, pages 2508--2516. IEEE, 2013.

\end{thebibliography}
}

\end{document}